\documentclass{IOS-Book-Article}

\usepackage{mathptmx}
\usepackage{amsmath}
\usepackage[utf8]{inputenc}   
\usepackage[T1]{fontenc}
\usepackage{textcomp}
\usepackage{url}

\usepackage[english]{babel}
\usepackage[autostyle, english = american]{csquotes}
\MakeOuterQuote{"}

\usepackage{graphicx}
\usepackage{booktabs}
\begin{document}
\begin{frontmatter}              

\title{That's So FETCH: Fashioning Ensemble Techniques for LLM Classification in Civil Legal Intake and Referral\\ }
\runningtitle{That's So FETCH}

\author[A]{\fnms{Quinten} \snm{Steenhuis}%
\thanks{Corresponding Author: Quinten Steenhuis, Suffolk University Law School, 120 Tremont St, Boston, Massachusetts, USA; E-mail:
qsteenhuis@suffolk.edu.}}

\runningauthor{Q. Steenhuis}
\address[A]{Suffolk University Law School, Boston, MA}

\begin{abstract}
Each year millions of people seek help for their legal problems by calling a legal aid program hotline, walking into a legal aid office, or using a lawyer referral service. The first step to match them to the right help is to identify the legal problem the applicant is experiencing. Misdirected applicants may miss a deadline, experience physical abuse, lose housing or lose custody of children while waiting to connect to the right legal help. We introduce and evaluate the FETCH classifier for legal issue classification and describe two methods for improving accuracy: a hybrid LLM/ML ensemble classification method, and the automatic generation of follow-up questions to enrich the initial problem narrative. We employ a novel dataset of 419 real-world queries to a nonprofit lawyer referral service. We achieve 97.37\% accuracy (hits@2) with inexpensive models, slightly above GPT-5, while reducing cost of guiding users to the right legal resource.
\end{abstract}

\end{frontmatter}

\thispagestyle{empty}
\pagestyle{empty}

\section*{Introduction}
Finding the right lawyer can be hard. Even those who know a lawyer may find they practice in the wrong area, while low-income applicants struggle to identify the right aid program and others must rely on friends or referral services. Getting legal triage or referral wrong means that the applicant may lose access to timely legal help and risk liberty, physical abuse or injury, loss of housing, or custody of a child.

One tool for classifying natural language text is a large language model (LLM). Even current state-of-the-art LLMs do not classify with 100\% accuracy. "Ensemble" classification, which combines multiple classification methods to produce a single score \cite{Breiman_1996_bagging}, is a method of improving classification accuracy.

This paper introduces the FETCH (Fast Ensemble Tagging and Classification Helper) classifier. FETCH is a multi-label weighted ensemble voting classifier that combines LLMs, keyword matching, and traditional ML.

This work was produced in collaboration with the Virginia Legal Aid Society and the Oregon State Bar. Virginia Legal Society (VLAS) receives approximately 18,000 calls to human intake workers each year. The Oregon State Bar (OSB) uses trained staff to field approximately 100,000 inquiries to its bar referral service each year.

Ultimately, we show that a combination of three inexpensive LLM models: GPT-5-nano, gemini-2.5-flash, and mistral-small, together with keywords and the traditional ML Spot API \cite{colarusso2022spot} may meet or exceed the performance of the current state-of-the-art GPT-5 model at a significant cost and speed savings.

\subsection{Research questions}
\begin{itemize}

    \item RQ1: Can small-model ensembles deliver equivalent or superior classification accuracy to frontier models while reducing cost in legal intake?
    \item RQ2: What types of legal intake cases remain most challenging for ensemble and LLM classifiers, and how can error analysis inform classifier design? 
\end{itemize}

\section{Prior work}
\subsection{AI for access to justice}
The application of AI to the domain of access to justice is well established and is now the topic of a biannual workshop co-located with JURIX and ICAIL \cite{ai4a2j2025}, as discussed in \cite{SteenhuisWestermannGettingInDoor2024}. See, e.g., applications to legal information tools \cite{branting2001advisory, westermann_justicebot_2023}, form-filling and expert systems \cite{steenhuis_digital_2021,  steenhuis2024ai, westermann2024dallma}, and dispute resolution \cite{ Westermann_thesis_2023, bickel2015online, Branting2022-BRAACM-5}.

\subsection{Ensemble classification of natural language text}
Ensembles have been used to improve accuracy when classifying text for decades \cite{Breiman_1996_bagging, opitz_popular_1999}, and LLMs have extended ensemble techniques to medicine \cite{qiu_ensemble_2025} and law \cite{FarrLLMChainEnsemble, ensemble_priyadarshini2021ledocl, sakai2025quadllmmltclargelanguagemodels}. 

\subsection{Legal intake and advice}
\cite{SteenhuisWestermannGettingInDoor2024} applied LLMs to legal intake, specifically threshold eligibility determinations in the legal aid context. The work in this paper builds on and generalizes that prior work.

\section{The legal problem classification task}
An attempted legal referral can end in one of three ways:

\begin{itemize}
    \item The applicant is matched with a provider who specializes in the kind of problem the applicant is experiencing.
    \item The intake worker may decide the applicant has a problem that does not have a legal solution, due to inapplicability, statutes of limitation, or jurisdiction.
    \item The intake worker may decide that the applicant is not experiencing a legal problem at all, but instead may be under a delusion.\footnote{A classic, commonly experienced example described by staff at the Oregon State Bar was "I have a microchip in my head". The phenomenon of body-control delusions has a well-documented link to paranoid schizophrenia dating to 1919. See, e.g., \cite{Tausk1933Microchip}.} In these cases, the intake worker may pay extra attention to handle the applicant's problem with sensitivity.
\end{itemize}

Classification is central to each of these tasks. Applicants may have an idea of the kind of lawyer they need to help with their problem, but that idea may be wrong.\footnote{For example, a layperson might be asked to decide that their problem getting Social Security Disability benefits falls under the category "Administrative Law" (a relatively obscure term for a layperson).} Informally, we observed applicants struggle when faced with distinguishing between abstract legal categories, and this is supported by studies such as \cite{world_justice_project_global_2019} (showing that only 29\% of people experiencing a legal problem recognize it as such) and \cite{oecd2019legalneeds} (emphasizing that self-assessed legal understanding often diverges from actual understanding).

Perfect referral is not always possible in a given geographical area, and applicants may settle for a next-best match in those cases.

\section{The FETCH classifier}
The FETCH classifier is available as a REST-like API. The API is in use by two separate applications: the Oregon State Bar's online lawyer referral service, and the Virginia Legal Aid Society's phone-based intake system.

We employ a single prompt across three LLMs.\footnote{https://gist.github.com/nonprofittechy/ef36fb8da928f25c60cd0ecb82a80750} The prompt includes the taxonomy and the applicant's problem description to determine if:

\begin{enumerate}
    \item The problem can be classified into one or more of 244 taxonomy nodes.
    \item More information is needed to accurately classify the applicant's problem.
    \item The applicant has not described a legal problem at all.
\end{enumerate}

We experimented with different levels of detail in the taxonomy for the prompt, but settled on an abbreviated level, with simply the name of each category and its parent category. Instructions were adjusted iteratively to balance generating follow-up questions against making a best effort to classify the applicant's problem.

The results are then scored by the number of times each node appears in the results of one of the models. The models are manually weighted according to performance on a smaller 20-query subset of the dataset and results are ranked accordingly. Finally, the top two nodes are shown to the applicant.

If the ensemble does not have sufficient confidence in the classification, each LLM model in the ensemble is instructed to generate 3 follow-up question candidates. The top 3 questions (after semantic merging with the assistance of GPT-5-nano) are then shown to the applicant.

\section{Oregon State Bar's referral form}

\begin{figure}
    \centering
    \includegraphics[width=0.5\linewidth]{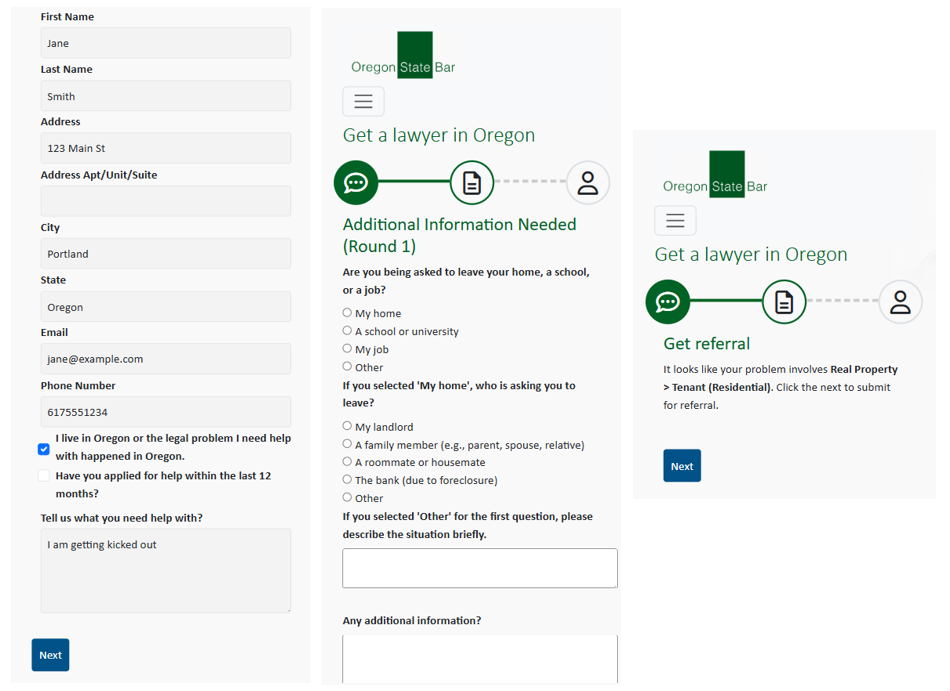}
    \caption{Screenshot of the lawyer referral service with integrated FETCH classifier, showing an automatically generated follow-up information form. Here, the applicant has entered an ambiguous query: "I am getting kicked out" and the FETCH classifier generates 3 questions to clarify if this is an eviction or other civil matter.}
    \label{fig:OSB-screenshots}
\end{figure}

The first implementation of the FETCH classifier is in the Oregon State Bar's online referral system (Figure \ref{fig:OSB-screenshots}). As shown in Fig. \ref{fig:OSB-screenshots}, applicants interact with the system through a consistent form metaphor, rather than in the form of a chatbot. We hypothesize that the form both reduces effort for the applicants and reduces the "uncanny valley" \cite{wang_uncanny_2015} effect that users experience when talking to an LLM \cite{radivojevic_human_2024}.

Because it generates follow-up questions automatically, the FETCH classifier enables the tool to gather more information without a human having to call the applicant.

Applicants who are screened out are given the option to call a human intake worker at the Oregon State Bar.

\section{Experimental design}

\subsection{Dataset}
In our experiment, we employed a novel dataset of 419 human-annotated queries made to the Oregon State Bar Referral Service over a period of a few weeks in 2025. The dataset contains a total of 31,716 words, with a mean problem description length of 74 words. Approximately 300 of the queries were annotated to one of the 15 top levels of the legal taxonomy only, while the 119 remaining queries were annotated to a specific "terminal node" in the OSB taxonomy. Annotation was performed by legal professionals at the Oregon State Bar during the regular operation of their referral program. The dataset was anonymized by the bar referral service prior to our analysis.\footnote{A small subset with additional deidentification is available publicly at https://gist.github.com/nonprofittechy/2aa8bfc7141569c7fdd7591564d8e270}

\begin{figure}
    \centering
    \includegraphics[width=0.5\linewidth]{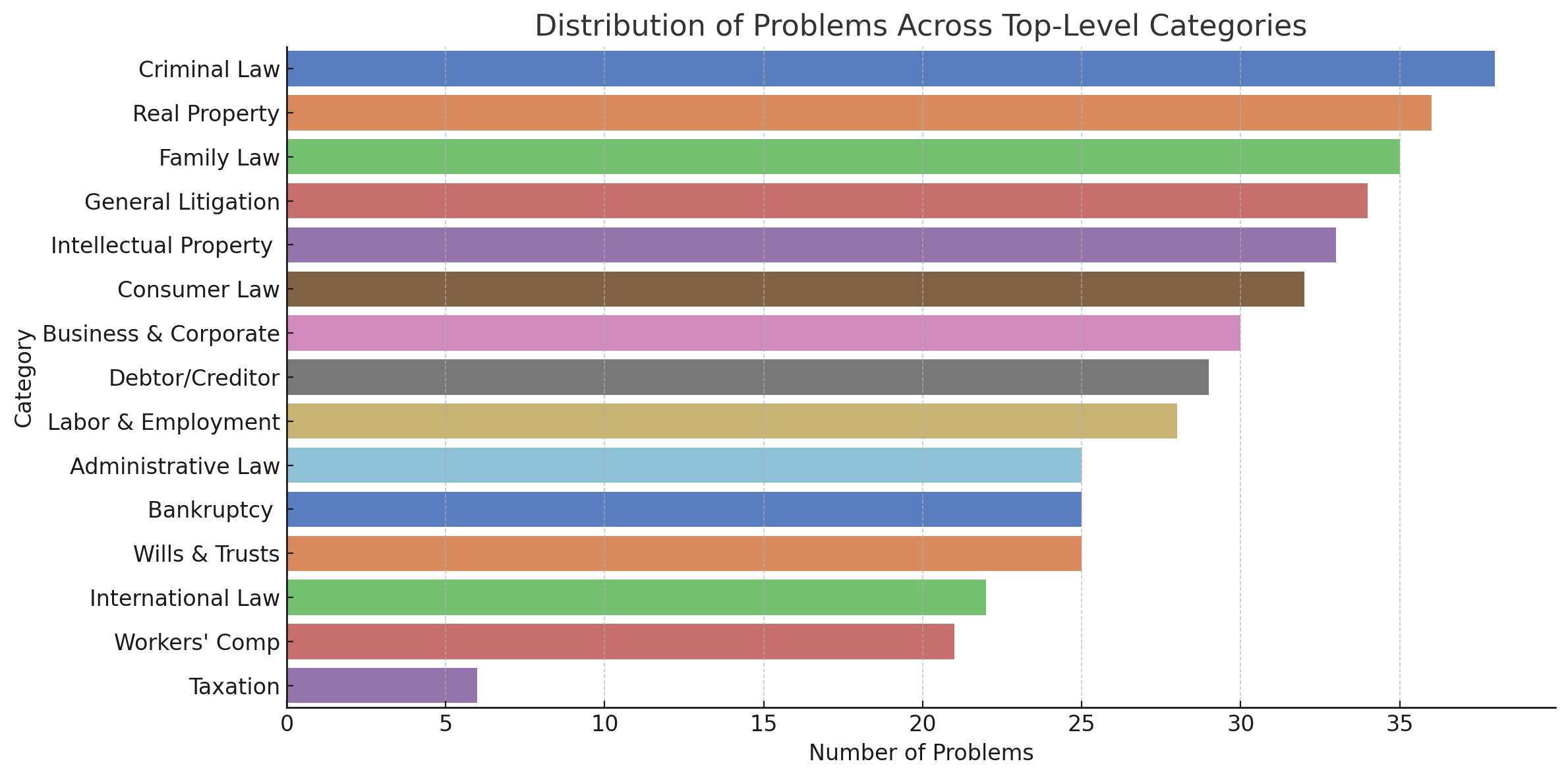}
    \caption{Distribution of problem type by top-level category name in the dataset.}
    \label{fig:enter-label}
\end{figure}

\begin{figure}
    \centering
    \includegraphics[width=0.5\linewidth]{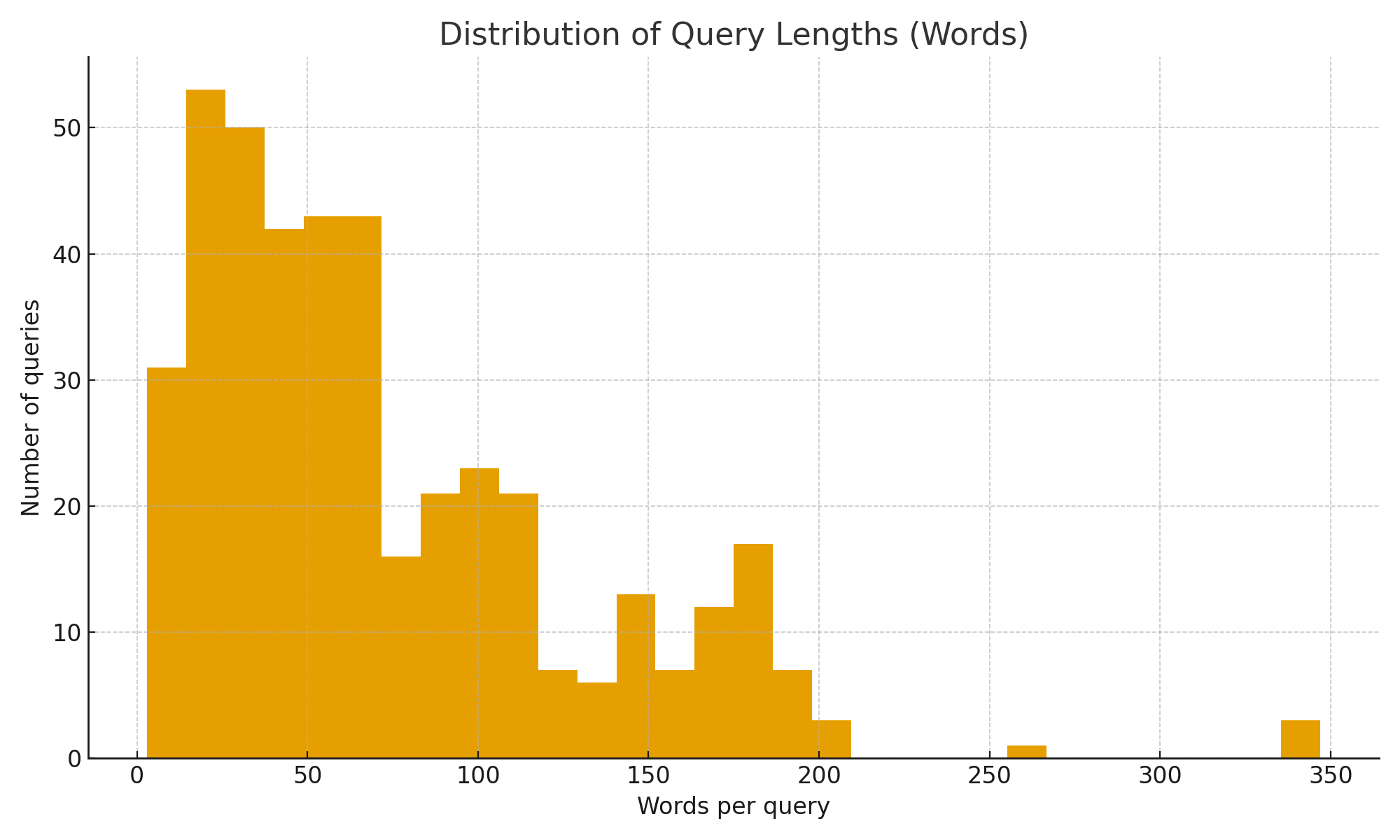}
    \caption{Distribution of applicant query lengths in the dataset.}
    \label{fig:query-length}
\end{figure}

Queries in the dataset show a diversity in length (Fig. \ref{fig:query-length}) and detail. Spelling errors and idiosyncratic punctuation appear throughout.

Examples ranged from short and direct:
\begin{itemize}
    \item Need bankruptcy lawyer
    \item Major car accident - DUII

To high-stakes problems, from stalking and domestic violence abuse to emergency housing (details redacted for length and privacy):\footnote{Unlike in \cite{SteenhuisWestermannGettingInDoor2024}, we did not observe any instances of model "censorship" or refusal to classify situations involving violence or abuse, a promising indication of the suitability of LLMs for problems in the legal domain.}

    \item I was granted a temporary stalking order on [...]. A contested hearing is scheduled for [...] and I need an attorney [...]. I would also like further legal advice on taking legal action against someone who is stalking me. [...] Months after I had stopped working with her as a housing advocate, she [...]
\end{itemize}

\subsection{Taxonomy}
Our study uses a 244 node taxonomy maintained by the Oregon State Bar,\footnote{For the full taxonomy, see https://gist.github.com/nonprofittechy/c8a85248d3835d77193f88d1ff41eda3} but national in its application. This represents a superset of alternative taxonomies such as the 135-node Legal Matter Specification Standard (commonly referred to as SALI taxonomy) \cite{sali2025lmsstaxonomy} and the smaller civil legal aid-focused LIST Taxonomy.\footnote{https://taxonomy.legal/}

\subsection{Classifiers}
Our study encompasses 9 separate classification approaches:

\begin{itemize}
    \item A simple keyword-based classifier.
    \item A term frequency/inverse document frequency classifier.
    \item GPT-4.1-mini and GPT-4.1-nano.
    \item Gemini-2.5-flash.
    \item Mistral-small-3.2-24b-instruct.
    \item Suffolk Legal Innovation and Technology Lab's traditional ML Spot \cite{colarusso2022spot} classifier (build 10).
    \item GPT-5 and GPT-5-nano, released near the end of our study.
\end{itemize}

The ensemble method was a simple weighted vote, with weights calculated based on the raw performance of each classifier across a subset of the training dataset. The classifier was limited to providing no more than the 2 "best" labels based on the weighted ranking.

We are able to use the additional signal of a confidence level from the SPOT classifier. For Gemini models, we use logprobs as returned in the LLM's response as a proxy for confidence level, as suggested by \cite{FarrLLMChainEnsemble}.\footnote{Unfortunately, the GPT-5 model series from OpenAI no longer makes logprobs available.}

\subsection{Evaluation method}
We evaluated the raw performance of the classifier with the help of the open-source Promptfoo tool against the human-annotated dataset with a hits@2 scoring system. We scored a result as "pass" if the human-annotated label was present in the response, and scored it as passing with a score of 0.5 if at least the top-level node was present in the labels.

\section{Results}
\subsection{Hits@2 results}
Table \ref{results-table} shows the results as scored using hits@2. The keyword classifier set a baseline at 54\%, while TF-IDF fared worst. Notably, the results show that the ensemble method performs best, meeting or slightly exceeding the performance of best-of-class GPT-5 alone\footnote{While suggestive, the .0.71 margin between GPT-5 and the ensemble is likely to be statistically insignificant.} and greatly exceeding the average performance of each inexpensive model used in the ensemble classifier.
\begin{table}[ht]
\centering
\small
\begin{tabular}{l r}
\toprule
Method & Hits@K (top 2 labels) (\%)\\
\midrule
Spot \cite{colarusso2022spot} & 59.43 \\
Keyword & 54.18 \\
TF-IDF& 31.03 \\
Gemini 2.5-flash & 87.35 \\
Mistral small & 87.83 \\
GPT-4.1-mini & 81.38 \\
GPT-5-nano & 87.11\\
GPT-5 & 96.66 \\
Ensemble (GPT-5-nano, Gemini 2.5-flash, Mistral small, Keyword, Spot) & \textbf{97.37} \\
\bottomrule
\end{tabular}
\caption{Model and ensemble accuracies.}
\label{results-table}
\end{table}

\subsection{Cost efficiency}
An important justification for the use of weighted ensemble voting is cost efficiency. As shown in Table \ref{cost-table}, the ensemble voting method outperforms the state-of-the-art GPT-5 model on a cost basis by a factor of more than 3.

\begin{table}[ht]
\centering
\small
\begin{tabular}{lrrrrr}
\toprule
Model & Cached & Uncached & Output & Avg. cost& Est. annual\\
      & (\$/1k) & (\$/1k) & (\$/1k) & (\$/1k req) & cost 100k (\$) \\
\midrule
mistral-small-3.2-24b-instruct & 0.026 & 0.008 & 0.030 & 0.064 & 6.40 \\
gemini-2.5-flash               & 0.039 & 0.048 & 0.750 & 0.837 & 83.75 \\
GPT-5                          & 0.065 & 0.201 & 3.002 & 3.267 & 326.71 \\
GPT-5-nano                     & 0.003 & 0.008 & 0.120 & 0.131 & 13.07 \\
GPT-4.1-nano                   & 0.013 & 0.016 & 0.120 & 0.149 & 14.91 \\
GPT-4.1-mini                   & 0.052 & 0.064 & 0.480 & 0.596 & 59.64 \\
\bottomrule
\end{tabular}
\caption{Cost model assumes a typical query of \textbf{519.5 cached input}, \textbf{160.5 uncached input}, and \textbf{300.2 output} tokens. Provider assumptions: OpenAI cached input is priced at \textbf{10\% of the input rate}; Gemini 2.5 Flash cached input is \textbf{\$0.075 per 1M tokens}; OpenRouter \texttt{mistral-small-3.2-24b-instruct} treats cached input at the \textbf{same price as input}. Note that the Spot API is free for nonprofits.}
\label{cost-table}
\end{table}

The latency of the small LLMs is also much lower. For example, GPT-5 has a minimum latency of over 5 seconds \cite{gpt5api} compared to GPT-5-nano's minimum latency of 2.2s \cite{gpt5nanoapi}.

\subsection{Errors}
We closely examined the 11 errors of the ensemble classification method. As in \cite{SteenhuisWestermannGettingInDoor2024}, we discovered probable human annotation errors through the LLM classification, totaling four such errors. We also discovered two duplicate entries (omitted in Table \ref{results-table}) in the human annotated dataset which may overweight the LLM classification errors.\footnote{We leave these annotation errors in our tops@2 score for consistency.} The full error results are summarized in Table \ref{results-table}. 

\begin{table}[ht]
\centering
\small

\begin{tabular}{@{}p{0.34\textwidth} p{0.23\textwidth} p{0.26\textwidth} p{0.17\textwidth}@{}}
\footnotesize \textbf{Short summary} & \textbf{Human annotation}& \textbf{LLM}& \textbf{Error classification}\\\hline
Roofing replacement contract; contractor obligations/next steps & Realty > Construction & General, Consumer & LLM error\\
Neighbor harassment/stalking (incl. bear spray); safety concerns & Realty > Neighbor Disputes & General > Neighbor Disputes/, General > Personal Injury & Synonymous\\
Dispute over dog ownership/possession with ex-partner; access withheld & Consumer  & General > Animal , Family > Divorce & Ambiguous\\
Wedding planner using event photos/video for promotion; cease-and-desist requested & Consumer  & General > Torts/Privacy, Intellectual Property > Trademark & Ambiguous\\
Lost trial for mobile-device while driving; seeks appeal & Consumer  & Criminal  > Appeals, Criminal  > Traffic Offenses & Human error\\
Wants help filing a small-claims case & Criminal  & Consumer  > Small Claims Advice, General& Human error\\
Car accident; likely uninsured; facing costs; asks about payment plans & Debtor/Creditor & General > Personal Injury, General > Property Damage & LLM error\\
Pandemic unemployment benefits issue; denial/overpayment and appeal options & Debtor/Creditor & Administrative  > Unemployment, Labor \& Employment > General & Human error\\
Physician Assistant practice without supervising MD/DO; licensing/supervision compliance & Labor \& Employment & Administrative  > Professional Licensing, Business > General & Human error\\
\end{tabular}
\caption{Detailed error results.}

\end{table}
For the two errors we classified as clear "LLM error," the LLM response was a form of "general litigation" instead of a more specific category. FETCH can optionally generate follow-up questions, and in four of the error results it did generate apparently relevant questions. We therefore examined how these automatically generated questions perform in practice.

\subsection{Follow-up questions for ambiguous queries}
Many real-world applicant queries are short, vague, or ambiguous. In other cases, while the applicant may describe a situation with multiple legal issues, they may only be able to find an attorney to help with one. Automatic follow-up questions can help clarify the applicant's intent and make legal distinctions that the applicant may not be aware of.
\subsubsection{Methodology}
The FETCH classifier generates up to three follow-up questions when confidence is low. The applicant's response is then submitted for an enriched attempt at classification.

To qualitatively test the value of this follow-up question mechanism, we selected 3 candidate questions from the dataset and treated them as follows: one question had details artificially removed. A second question was unchanged, but had been classified by OSB as someone without a legal problem. A third question was misclassified by the LLM. Then we reviewed the output of FETCH for relevance and helpfulness of the follow-up questions to reach an appropriate classification. The results are analyzed in Table \ref{tab:followup-examples}
\begin{table}[ht]
\centering
\renewcommand{\arraystretch}{1.2} 
\begin{tabular}{|p{0.2\textwidth}|p{0.35\textwidth}|p{0.3\textwidth}|}
\hline
\textbf{Query}& \textbf{Generated Follow-up Questions} & \textbf{Analysis} \\
\hline
I need help with my baby. There’s a domestic situation &
1. Is this about the baby's parents or guardians who are or may be divorcing or separating? \newline
2. Is there safety risk now for you or your baby, or are you worried about safety? (Options: Yes, there is immediate danger, There is some risk but not immediate, No risk currently, Not sure) \newline
3. Do you need help with child custody or visitation for your baby? (Options: Yes, No, Not sure) &
(Context removed). The generated questions identify the two potential legal issues: divorce and restraining order, and a correct answer would match the original classification.\\
\hline
I have degenerative arthritis in my ankle and have to wear a brace, making it hard to walk unless I use a walker &
1. Are you seeking financial assistance or benefits due to your condition? \newline
2. Has your condition affected your ability to work or keep a job? \newline
3. Are you experiencing any problems with accessibility in public places or at your home because of your difficulty walking? &
(full question, no attorney assigned by OSB). The clarifying questions are relevant and identify potential legal issues, such as denial of benefits or Americans with Disabilities Act compliance.\\
\hline
There was a car accident [...] I saw a letter, explaining I owe \$10{,}000 due to the car crash, [...] if any payment plans could be done and I need a lawyer for legal advice. &
1. What is the current status of this case? (Options: Lawsuit filed or court filing, Demand letter or collection letter only, Not sure/unknown)\newline
2. Were you at fault in the car accident? (Options: Yes, No, Unsure)\newline
3. Was there a police report about the accident?&
(full question, misclassified by LLM as dealing with the car crash) The clarifying questions do get us closer to the human annotator's decision that despite originating with a car accident, at core this is a debtor/creditor problem.\\
\hline
\end{tabular}
\caption{Examples of ambiguous applicant queries and clarifying follow-up questions generated by FETCH.}
\label{tab:followup-examples}
\end{table}

\section{Discussion}

\subsection{RQ1: Can small-model ensembles deliver equivalent or superior classification accuracy to frontier models while reducing cost in legal intake?}
We have shown that small-model ensembles substantially improve on the performance of older machine learning and keyword-matching approaches to classifying legal problems, and meet or slightly improve on the performance of frontier LLM models. What's more, they do so with lower latency and at a significantly lower cost, at least $\tfrac{1}{3}$ of the cost of using a frontier model for the same task. There was no evidence of bias in the distribution of errors and safety concerns revealed by the errors appear to be minimal.

The LLM-enabled classification approach shows promise to offer nonprofit bar referral services and legal aid programs a cost-effective way to deliver high quality referral or eligibility determination information to applicants for legal help.

\subsection{RQ2: What types of legal intake cases remain most challenging for ensemble and LLM classifiers, and how can error analysis inform classifier design? }
We did not observe any patterns in the errors in our ensemble classifier's results. They did not appear to be biased towards one type of legal problem over another, and similar categories were not consistently confused by the ensemble. Importantly, this tells us that enriching the category descriptions or adding additional clarifying instructions in the prompt is not likely to significantly improve classification performance.

Overall, annotation errors and ambiguities suggest human baseline performance may match the ensemble. Initial qualitative results suggest the follow-up questions would further improve the performance of the system in real-world usage.

Overall, the errors made by the LLM appear to be remarkably few (approximately 2\%), and consistent with the kind of challenges humans may face in similar circumstances.

\subsection{Limitations and risks}
The results of the study are strong. However, there are some limitations:
\begin{itemize}
    \item While a relatively large dataset in this domain, 419 queries may not adequately cover the 244 nodes in the full taxonomy. Our findings are strongest when considering the 15 top-level categories.
    \item During our study, OSB was only able to collect 19 "no legal problem" scenarios, not enough for a robust analysis. Therefore we did not fully test the "no legal problem" condition, especially delusional queries, which will require future sensitivity testing.
    \item We did not measure whether faster automated referral improved real outcomes.
    \item While bias is important to consider when using LLMs \cite{bender_stochastic_parrot}, the low error rate, as well as the diversity of methods and models in the ensemble, reduces this concern. Built-in human review is an additional safeguard in this system.
\end{itemize}
 
\section{Conclusion and future work}
The ensemble method shows promise to improve performance on a high-stakes classification task. Importantly, our results show that a weighted combination of inexpensive small language models can slightly outperform the current state-of-the-art model, GPT-5, at approximately $\tfrac{1}{3}$ of the cost.

Although the error rate is small in our study, future work could further narrow the gap. A routing method based on perceived confidence or disagreement among the model responses could supplement the inexpensive LLM models with a call to a best-of-class LLM such as GPT-5. We additionally hope to run a broader, quantitative study of the value of follow-up answers in improving classification. One approach would be to manually remove information from real queries. Gathering information about  the baseline human error rate would add additional value.

\section*{Acknowledgments}
    The author thanks Virginia Legal Aid Society and Oregon State Bar for funding and data access and David Colarusso, Jack Adamson, and Jim Graszer for their contributions and support of this project.

\bibliographystyle{acm}
\bibliography{bibliography, opening_door_bib}

\end{document}